\documentclass{article}
\usepackage{spconf,amsmath,amssymb,graphicx,subfigure}
\usepackage{epstopdf}
\usepackage{algorithmic, algorithm,color,multirow,eqnarray}
\usepackage[space]{grffile}
\usepackage{cite}

\usepackage{fancyheadings}
\title{Evolutionary Projection Selection \\ for Radon Barcodes}
\name{H.R. Tizhoosh$^\dagger$, S. Rahnamayan$^\ddagger$}
\address{$^\dagger$ KIMIA Lab, University of Waterloo, Waterloo, ON\\Canada  N2L 3G1, 
\texttt{tizhoosh@uwaterloo.ca} \\
$^\ddagger$ Electrical, Computer and Software Engineering, University of Ontario Institute of Technology\\ Oshawa, Ontario, Canada, \texttt{shahryar.rahnamayan@uoit.ca}}
\begin{document}
\maketitle
\begin{abstract}
Recently, Radon transformation has been used to generate barcodes for tagging medical images. The under-sampled image is projected in certain directions, and each projection is binarized using a local threshold. The concatenation of the thresholded projections creates a barcode that can be used for tagging or annotating medical images. A small number of equidistant projections, e.g., 4 or 8, is generally used to generate short barcodes. However, due to the diverse nature of digital images, and since we are only working with a small number of projections (to keep the barcode short), taking equidistant projections may not be the best course of action. In this paper, we proposed to find $n$ optimal projections, whereas $n\!<\!180$, in order to increase the expressiveness of Radon barcodes. We show examples for the exhaustive search for the simple case when we attempt to find 4 best projections out of 16 equidistant projections and compare it with the evolutionary approach in order to establish the benefit of the latter when operating on a small population size as in the case of micro-DE. We randomly selected 10 different classes from IRMA dataset (14,400 x-ray images in 58 classes) and further randomly selected 5 images per class for our tests.        
\end{abstract}

\section{Introduction}
Searching in large databases and archives of digital images to find duplicates or similar instances of a given input (query) image is a difficult task, both with respect to desired accuracy and affordable computational power. Generally, we may use various search engines to retrieve images when we provide a set of keywords that describe what we like to see. However, the text-based image search is very limited in its performance. You easily find suitable candidates if you search for ``red cars''. It becomes more difficult if you look for ``red cars with a flat front tire'' or ``red cars whose door handle is broken and the front tire is flat''. You may be lucky and find some cases by accident. Such cases are embedded in web pages that do contain some of the keywords you used to describe what you are looking for.

Text-based search for images, however, has many problems. Not all images are annotated with meaningful text, and not all components of a digital image can be adequately and uniquely described with words. Satellite images, x-ray images, histopathology images are examples for this class of digital images where the image content cannot be captured in textual format. 

Searching for images based on what they contain is called ``content-based image retrieval'' (CBIR) \cite{rajam2013,ghosh2011}. The content of the image, of course, is comprised of pixels, colors, edges, textures, and segments. Describing the image content with a focus on any of these ``features'' can guide CBIR in a very different direction. One of the recent trends in computer vision community is to generate and use binary descriptors such as Local Binary Patterns (LBP) \cite{ojala2002,ahonen2006}, Binary Robust Invariant Scalable Keypoints (BRISK) \cite{Leutenegger2011}, and Radon BarCodes (RBC) \cite{Tizhoosh2015,Tizhoosh2016}. In this paper, we focus on the latter to contribute to the generation of a more expressive Radon barcode for tagging medical images. Expressive projection angles are understood to be non-redundant projections that minimize the reconstruction error, hence contributing to more compact barcodes effective for accurate image search.    

The paper is organized as follows: Section \ref{sec:bkg} provides a brief review of the relevant literature. In section \ref{sec:rbc}, Radon barcodes are reviewed. Section \ref{sec:prop} describes the proposed approach to select a small number of projections that contribute to the generation of more expressive Radon barcodes. Experiments using images from IRMA dataset are described in section \ref{sec:exp}. The paper is summarized in section \ref{sec:conc}. 

\section{Background Review}
\label{sec:bkg}

The literature on CBIR is quite extensive \cite{ghosh2011,camlica2015autoencoding,shandilya2010,rajam2013,dharani2013}.
Although binary images have been used to facilitate image search in many different ways \cite{PatentConsensus,PatentQB,Daugman2004,Arvancheh2006,camlica2015medical,ojala2002}, it seems that no attempt has been made to binarize the Radon projections until recently \cite{Tizhoosh2015} to use them directly for image search. A binarized projection can serve as a descriptor for the image. The binary information is compact and fast for search. The main motivation for using Radon transform, and similar techniques, is depicting a three-dimensional object by building many projections around it, usually 180 projection angles to fully scan the object. There are many applications of Radon transform \cite{zhao2013,hoang2012,nacereddine2010,jadhav2009,daras2006}.  
 
Differential evolution (DE) is a simple and effective evolutionary algorithm proposed by Stron and Price in 1997 \cite{storn1997differential}. DE has three main control parameters, namely, population size, mutation scale factor, and crossover rate. During almost two decades, many variants and improved versions of DE have been proposed; the majority of them have tried to improve its convergence rate and robustness. The classical DE works directly on continuous search space, however, several discrete DE algorithm have also been introduced \cite{wang2010novel, pan2008discrete, tasgetiren2010ensemble}.  

The DE algorithm, similar to other population-based algorithms, suffers from a high computational cost. A large population size supports a higher exploration power while a small size reduces the number of function evaluations with the cost of less exploration capability and risk of premature convergence and stagnation. DE algorithm with a small population size (less than 10) is called micro-DE algorithm. Opposition-based learning (OBL) is a recently proposed new scheme in machine learning \cite{tizhoosh2005opposition, rahnamayan2008opposition}; this scheme is utilized to enhance the performance of the micro-DE algorithm for image thresholding \cite{rahnamayan2008image}. 

To tackle the stagnation and premature convergence problems in MDE algorithm, utilizing a vectorized random mutation (MDEVM) factor is proposed \cite{salehinejad2014micro}. The MDEVM generates a random mutation scale factor per dimension; the reported results are promising. Another method called $\mu$JADE \cite{brown2015mu}, is proposed recently; it is an adaptive differential evolution algorithm with a small population size. It uses a new mutation operator, called current-by-rand-to-pbest. Another research work used a pool of mutation strategies as well as a pool of values for each control parameter of the algorithm \cite{mallipeddi2011differential}. 

The ensemble algorithm is a recently proposed algorithm which uses a pool of mutation schemes with two members, DE/rand/1 and DE/best/1. The strategy is to combine random parameters setting and mutation schemes to enhance convergence rate while maintaining the diversity of population \cite{yu2015ensemble}. A switching approach is proposed in \cite{das2015switched}, which the scale factor and crossover rate are switched using a uniform random strategy in an individual level of population. A pool of mutation schemes is also utilized where the individuals use either the DE/rand/1 or DE/best/1 scheme. 

Utilizing a micro-DE algorithm in real-time or on-line applications seems reasonable, because of its low hardware resource requirement and low computational time characteristic. micro-DE algorithms are generally used for solving low dimensional optimization problems, which makes that a good candidate for solving our optimal projection problem raised in this paper. 
 
\section{Radon Barcode Generation}
\label{sec:rbc}
\begin{figure}[tb]
\begin{center}
\vspace{0.05in}
\includegraphics[width=0.80\columnwidth]{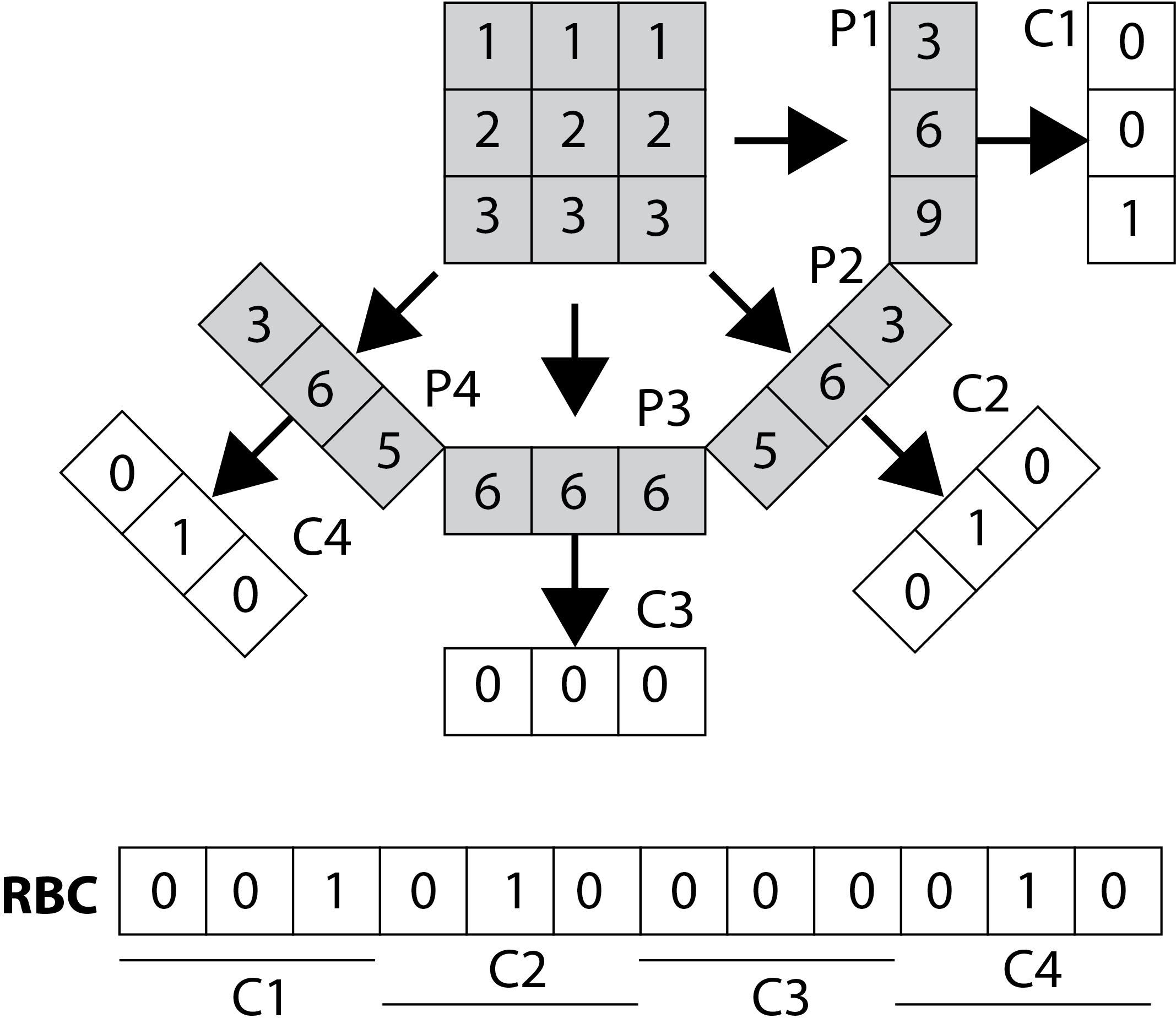}
\caption{Radon Barcode (RBC) \cite{Tizhoosh2015} -- Projections (P1,P2,P3,P4) are binarized (thresholded) to generate code fragments C1,C2,C3,C4. Putting all code fragments together delivers the barcode \textbf{RBC}. }
\label{fig:RBC}
\end{center}
\end{figure}

For any digital image described as a function $f(x,y)$, one can project $f(x,y)$ along a number of projection angles using Radon transform. The projection is basically the sum (integral) of $f(x,y)$ values along lines constituted by each angle $\theta$. The projection creates a new function $R(\rho,\theta)$ with $\rho = x \cos \theta + y \sin \theta$. Hence, using the Dirac delta function $\delta(\cdot)$ the Radon transform can be written as \cite{Radon1917,Tizhoosh2015}
\begin{equation}
R(\rho,\theta) = \int\limits_{-\infty}^{+\infty} \int\limits_{-\infty}^{+\infty} f(x,y) \delta(\rho-x\cos \theta-y\sin\theta) dx dy.
\end{equation}
In general, we use the projections to reconstruct the image (if projections are real scans of the physical object). In case of building the Radon transform of an existing image, we can threshold all projections (lines) for individual angles based on a ``local'' threshold for that angle, such that we can assemble a barcode of all thresholded projections as depicted in Figure \ref{fig:RBC}. A simple way for thresholding the projections is to calculate a typical value $T$ via median operator applied on all non-zero values of each projection (zero-padding is common to generate same-lentgh projection vectors for all angles). Algorithm \ref{alg:Radon} describes how \textbf{Radon barcodes (RBC)} can be calculated\footnote{Matlab code available online: http://tizhoosh.uwaterloo.ca/}.  In order to receive same-length barcodes \emph{Normalize$(I)$} downsamples all images into $R_N\times C_N$ images (i.e., $R_N= C_N$).

\begin{algorithm}[htbp]
\caption{Radon Barcode (RBC) Generation \cite{Tizhoosh2015}}
\begin{algorithmic}[1]
\label{alg:Radon}
\STATE Initialize Radon Barcode $\mathbf{r} \leftarrow \emptyset$ 
\STATE Set downsampling size $R_N=C_N\leftarrow 32,64,128,\dots$
\STATE Set number of projection angles $n_\theta\leftarrow 4,8,16,\dots$
\STATE Initialize angle $\theta \leftarrow 0$ 
\STATE Get the query image $I$
\STATE Downsample $I$:  $\bar{I} = \textrm{Normalize}(I,R_N,C_N)$ 
\WHILE{$\theta < 180$}
	\STATE Get all projections $\mathbf{p}$ for $\theta$
	\STATE Find typical value $T_\textrm{typical}\leftarrow\textrm{median}_i (\mathbf{p}_i)|_{\mathbf{p}_i \neq 0}$
	\STATE Binarize projections: $\mathbf{b} \leftarrow \mathbf{p} \geq T_\textrm{typical}$ 
	\STATE Append the new row $\mathbf{r} \leftarrow \textrm{append}(\mathbf{r},\mathbf{b} )$ 
	\STATE $\theta \leftarrow \theta + \frac{180}{n_\theta}$
\ENDWHILE
\STATE Return  $\mathbf{r}$
 \end{algorithmic}
 \end{algorithm}

\begin{figure}[htb]
\centering     
\subfigure[input image]{\label{fig:a}\includegraphics[width=30mm,height=30mm]{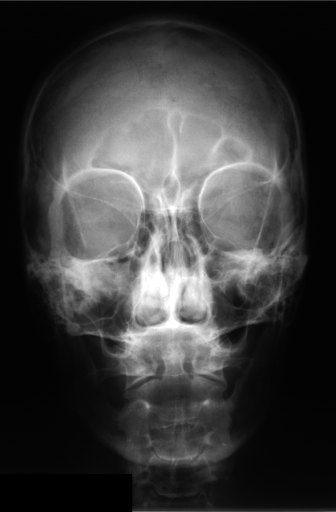}}\qquad
\subfigure[input image]{\label{fig:a}\includegraphics[width=30mm,height=30mm]{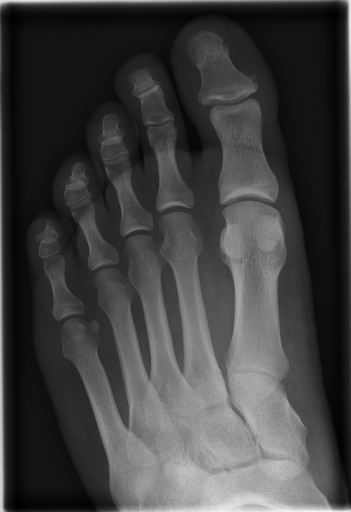}}\\

\subfigure[RBC$_8$]{\label{fig:a}\includegraphics[width=0.4\columnwidth,height=4mm]{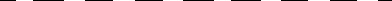}}\qquad
\subfigure[RBC$_8$]{\label{fig:a}\includegraphics[width=0.4\columnwidth,height=4mm]{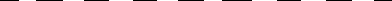}}\\

\subfigure[RBC$_{16}$]{\label{fig:a}\includegraphics[width=0.4\columnwidth,height=4mm]{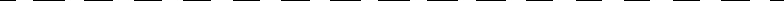}}\qquad
\subfigure[RBC$_{16}$]{\label{fig:a}\includegraphics[width=0.4\columnwidth,height=4mm]{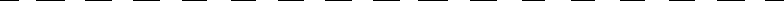}}\\

\subfigure[RBC$_{32}$]{\label{fig:a}\includegraphics[width=0.4\columnwidth,height=4mm]{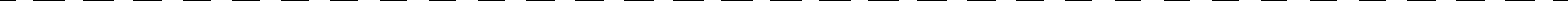}}\qquad
\subfigure[RBC$_{32}$]{\label{fig:a}\includegraphics[width=0.4\columnwidth,height=4mm]{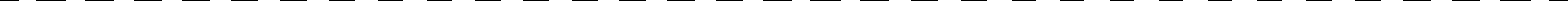}}\\

\caption{Radon Barcodes with 8/16/32 projection angles \cite{Tizhoosh2015}.}
\label{fig:Barcodes}
\end{figure}

Figure \ref{fig:Barcodes} shows barcode annotations for two medical images from IRMA dataset \cite{Lehmann2003,Lehmann2006} for different $n_\theta$ values (see lines 4 and 10 in Algorithm \ref{alg:Radon}). Hence, the algorithm works with $n_\theta<180$ equidistant angles of projection. 

\section{Proposed Approach}
\label{sec:prop}
As described in the previous section, the proposed Radon barcode is generated based on $n<180$ equidistance angles of projections. Of course, this may not be the best scheme for generating expressive binary descriptors for the content of different images may be ``seen'' from different angles with different levels of ease/difficulty (e.g., object occlusion occurs for specific angles). Hence, in this paper, we propose to use an evolutionary algorithm with a small population size to find the best selection for ${{180}\choose{n}}=\frac{180!}{n! (180-n)!}$. For $n=4$, we have to check 42,296,805 combinations by applying exhaustive search which is prohibitively expensive.      

Exhaustive search -- The most straightforward approach to finding $n<180$ optimal projections is to check all combinations ($n$ out of $180$). Of course, this is a problem with factorial complexity, $O(n!)$, and would be intractable in the practice. Instead, we may just test ($n$ out of $m$) whereas $m$ is a much smaller number, for instance, $m=8$ or $16$ equidistant projection angles. We did this experiment first for benchmarking purposes only. We established the optimal 4 projections out of 16 equidistant projections to have a reference for comparison. Of course, this one needs to check all 1820 possible combinations: ${{16}\choose{4}}=1820$.    

Evolutionary Search -- To be able to find $n$ optimal projection angles without getting into prohibitively long computational times, we can configure a fast evolutionary meta-heuristic like micro-DE \cite{rahnamayan2008image}. Basically, one attempts to find the solution with a small population size such that the evolutionary process is not sluggish. Assuming that micro-DE will do the job, the most important question is what is \emph{fitness function} to find the optimal projections for the Radon barcodes? 

\begin{figure}[tb]
\centering
\vspace{0.05in}
\includegraphics[width=0.8\columnwidth]{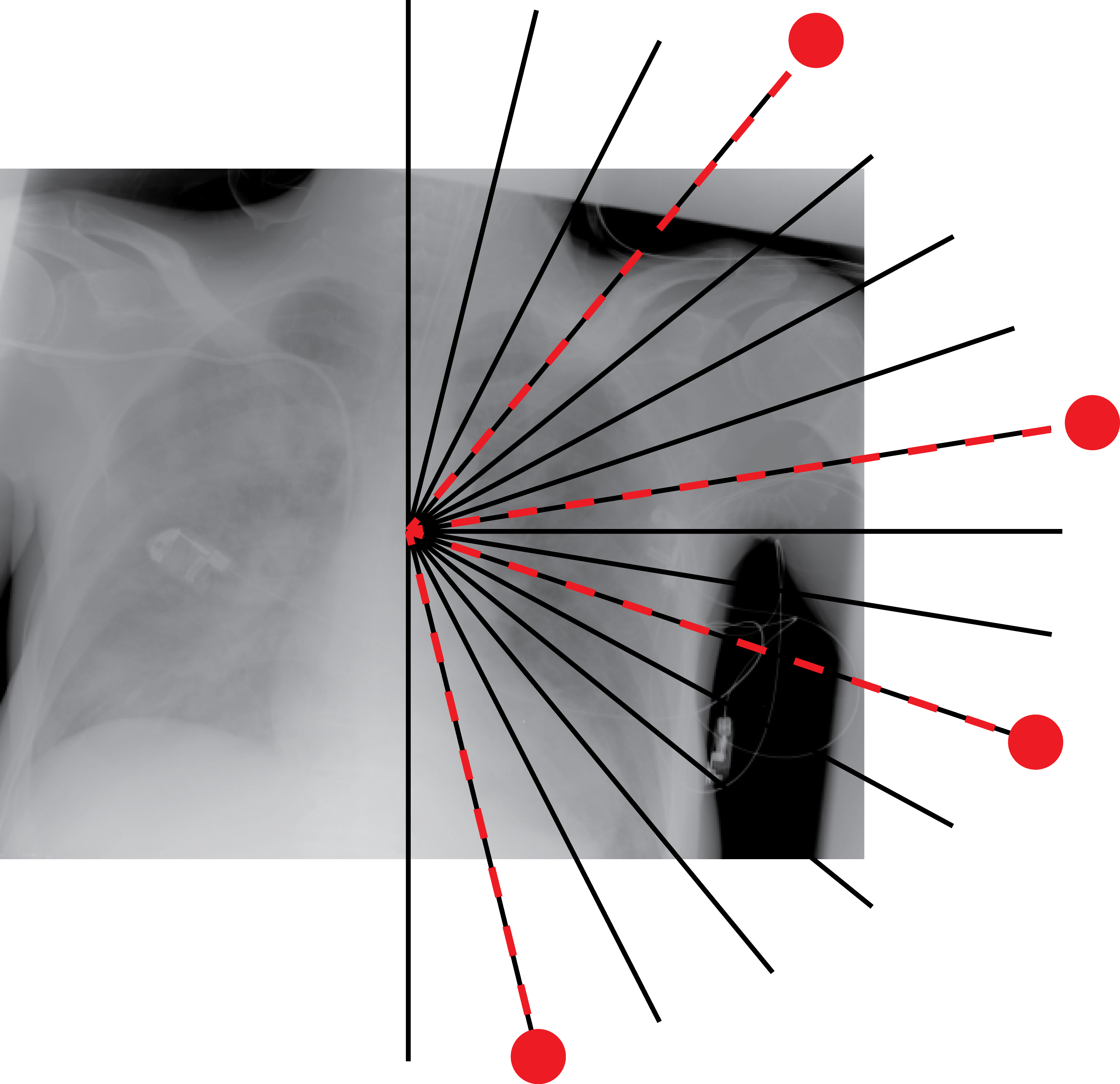}
\caption{4 Out of 16 equidistant projections, the four selected ones (here $33.75^\circ, 78.75^\circ, 123.75^\circ, 168.75^\circ$) via brute-force search in 1820 possible combinations,  minimize the reconstruction error if we apply the inverse Radon transform restricted to only using four projections out of 16 available. }
\label{fig:optimalproject}
\end{figure}

Radon barcodes are binarized versions of Radon projections. To create the optimal Radon barcodes (barcodes that are highly expressive to tag digital images in a unique way such that they can be retrieved easily), one has to find the optimal projections. Since we are trying to find $n<180$ projections, we should find a fitness function that guides micro-DE in the right direction. If the $n$ projections are good (compared to all other possible combinations), then image reconstruction using these $n$ projections should yield smaller error compared to all other combinations. Figure \ref{fig:optimalproject} shows for the chest x-ray that the 4 best projections by examining all 1820 combinations when 16 equidistant projections are available.

With $\rho$ being the slope and $\theta$ the intercept of a line (projection), Radon transform $R_{(\rho,\theta)}[f(x,y)]$ of the function $f(x,y)$ can be given as 
\begin{eqnarray}
R_{(\rho,\theta)}[f(x,y)] &=& \int\limits_{\infty}^{-\infty} f(x,\theta+\rho x)dx\\
 &=& \int\limits_{\infty}^{-\infty}\int\limits_{\infty}^{-\infty}f(x,y)\delta[y-(\theta+\rho x)] \\
&=& U(\rho,\theta)
\end{eqnarray}
The inverse Radon transform can be given using a Hilbert transform $H$ as follows 
\begin{equation}
f(x,y)=\frac{1}{2\pi} \int\limits_{\infty}^{-\infty} \frac{d}{dy} H[U(\rho,y-\rho x)] d\rho
\end{equation}
It is obvious that if we do not use all 180 projections, we will not be able to reconstruct $f(x,y)$ with low error but an imperfect version $\hat{f}(x,y)$ resulting from reconstruction by using $n_\theta<180$. Hence, we like to find $n_\theta$ projection angles such that
\begin{equation}
|f(x,y)-\hat{f}(x,y)|\rightarrow 0
\end{equation}

As this may not be feasible, we will use correlation between two $M\!\times\!N$ images that needs to be maximized (correlation here being a measure of image similarity): Find $n$ projections subject to correlation \emph{C}$(f,\hat{f})$ is maximum (with $\mu_f$ and $\mu_{\hat{f}}$ representing the mean values of each function): 
\begin{equation}
\label{eq:corr}
C\!=\!\frac{\sum\limits_{x=1}^{M}\sum\limits_{y=1}^{N}(f(x,y)-\mu_f)(\hat{f}(x,y)-\mu_{\hat{f}})}{\sqrt{\sum\limits_{x=1}^{M}\sum\limits_{y=1}^{N}(f(x,y)\!-\!\mu_f)^2\times  \sum\limits_{x=1}^{M}\sum\limits_{y=1}^{N}(\hat{f}(x,y)\!-\!\mu_{\hat{f}})^2}}
\end{equation}
    
Algorithm \ref{alg:MDERadon} describes the steps invloved in the proposed approach. The termination condition can be the maximum number of objective function calls (Eq.\ref{eq:corr}) or a  predefined value-to-reach.    

\begin{algorithm}[htbp]
\caption{General steps for using micro-DE for selection of optimal Radon projection angles. }
\begin{algorithmic}[1]
\label{alg:MDERadon}
\STATE Set the desired search space: $n_\theta$=4,8,16,$\dots$ out of 180 
\STATE Read the input image $f(x,y)$ 
\STATE Initialize the population for the micro-DE algorithm
\STATE Reconstruct $\hat{f}(x,y)$ with current best angles
\STATE Evaluate fitness value for all individuals using Eq. \ref{eq:corr} 
\WHILE{Termination condition not satisfied} 
\STATE Apply mutation scheme of DE algorithm (DE/rand/1)
\STATE Apply binary crossover 
\STATE Select the better candidates between the trial candidate and the parent individual
\ENDWHILE 
\STATE Return the best projections angles and corresponding correlation value.

\end{algorithmic}
\end{algorithm}

\section{Experiments}
\label{sec:exp}
\begin{figure}[htbp]
\centering
\includegraphics[width=0.99\columnwidth]{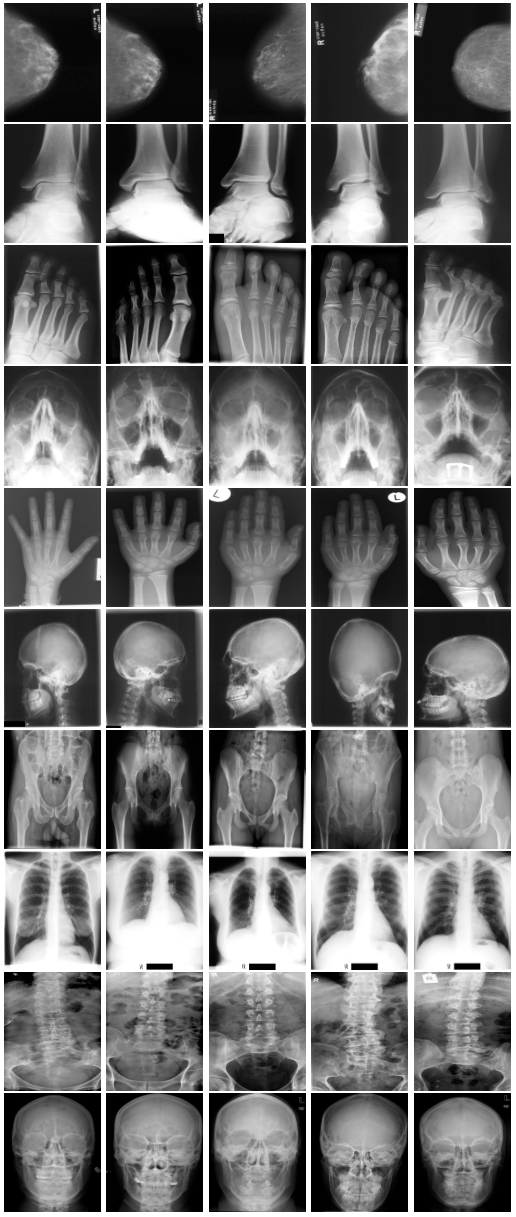} 
\caption{50 Randmly selected images from IRMA Dataset \cite{Lehmann2003,Lehmann2006} for our experiments. }
\label{fig:sampleXrays}
\end{figure}

 In this section, we report two series of experiments. The first one verifies the correctness or reliability of the micro-DE algorithm to find optimal projections compared to an exhaustive search when we are looking for 4 optimal projection angles out of 16 equidistant angles. In the second experiments, we still provide the results for exhaustive 4/16 selection but examine the evolutionary approach to get 4/180 and 8/180 to investigate the reconstruction accuracy of the micro-DE via correlation of the input image and the reconstructed image using 4/16 (exhaustive), 4/180 (micro-DE) and 8/180 (micro-DE). To conduct the experiments we used 50 images from IRMA dataset \cite{Lehmann2003,Lehmann2006}, a benchmarking collection of 14,400 x-rays images. We selected 10 random classes out of 58 classes, and for each class, we randomly drew 5 images to be used in our experiments. Figure \ref{fig:sampleXrays} shows the images we have used. 

\subsection{Experiment Series 1}

In order to verify the usability of the micro-DE for locating optimal projection angles to generate compact Radon barcodes, we need a baseline to compare against. Of course, this is a challenge because applying brute force to find the angles can generate the only reliable benchmark data. However, this is not feasible for the general case. We, therefore, reduce the problem of optimal angles for test images (4 out of 16) via exhaustive search 
(1820 combinations checked!). The results are illustrated in Table \ref{tab:BFMDE}. As it can be seen, not only micro-DE results are very close to the brute force results (close/same angles found) but also the correlation values are very close. This indicated that micro-DE is a reliable approach. 

\begin{table}[h]
\centering
\caption{Brute Force (BF) versus micro-DE (MDE): Results for four random images from IRMA Dataset \cite{Lehmann2003,Lehmann2006} to find 4 optimal angles out of 16 given equidistant ones. Correlation $C$ between the input image and the reconstructed image denotes the accuracy of image reconstruction.}
\label{tab:BFMDE}
\begin{tabular}{ll}
\hline
image & optimal 4/16 via BF and MDE \\ \hline  
\multirow{3}{*}{\includegraphics[width=1.5cm,height=1.5cm]{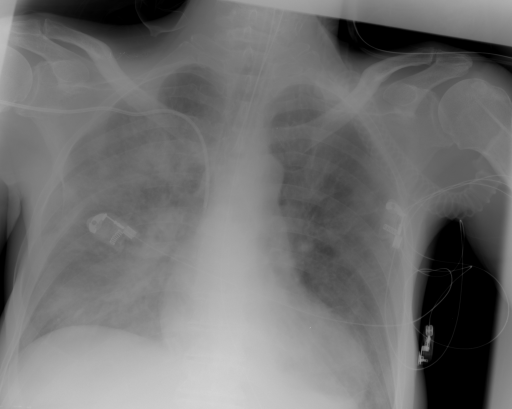}}  & 
	[34$^\circ$, 79$^\circ$, 124$^\circ$, 158$^\circ$] with $C=61\%$  \\
 & [45$^\circ$, 79$^\circ$, 124$^\circ$, 146$^\circ$] with $C=61\%$\\
  & \\
  & \\
\multirow{3}{*}{\includegraphics[width=1.5cm,height=1.5cm]{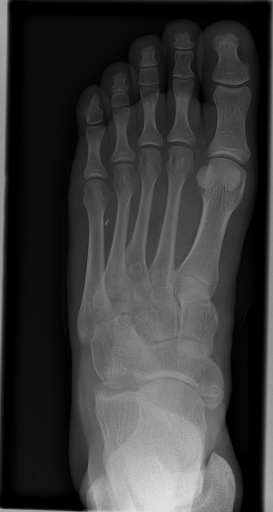}}  & 
	[22$^\circ$, 67$^\circ$, 112$^\circ$, 157$^\circ$] with $C=75\%$\\
 & [22$^\circ$, 56$^\circ$, 135$^\circ$, 169$^\circ$] with $C=74\%$ \\
 & \\
 & \\
\multirow{3}{*}{\includegraphics[width=1.5cm,height=1.5cm]{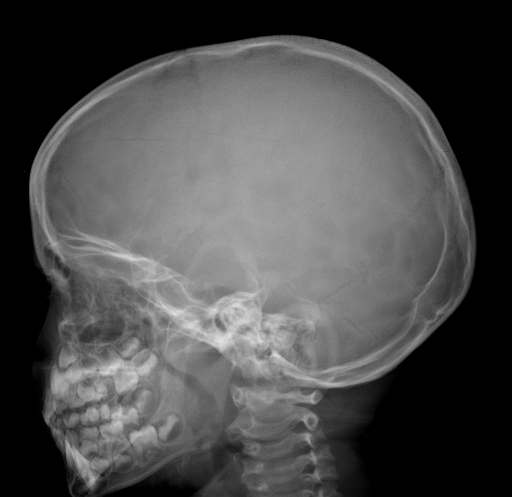}}  & 
	[0$^\circ$, 79$^\circ$, 101$^\circ$, 157$^\circ$] with $C=90\%$\\
 & [0$^\circ$, 68$^\circ$, 112$^\circ$, 169$^\circ$] with $C=89\%$\\
 & \\
 & \\
\multirow{3}{*}{\includegraphics[width=1.5cm,height=1.5cm]{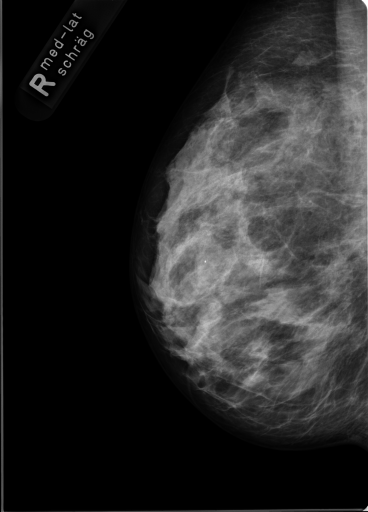}}  & 
	[22$^\circ$, 68$^\circ$, 124$^\circ$, 157$^\circ$] with $C=84\%$\\
 & [22$^\circ$, 68,$^\circ$,135,$^\circ$, 169$^\circ$] with $C=82\%$\\
 & \\
 & \\
\end{tabular}
\end{table}

\subsection{Experiment Series 2}
\begin{figure}[h]
\centering
\includegraphics[width=0.19\columnwidth,height=2cm]{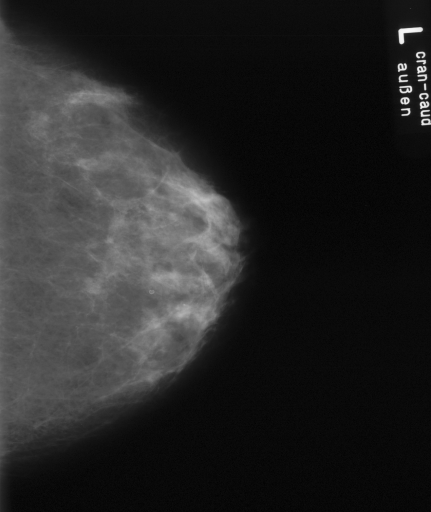}
\includegraphics[width=0.19\columnwidth,height=2cm]{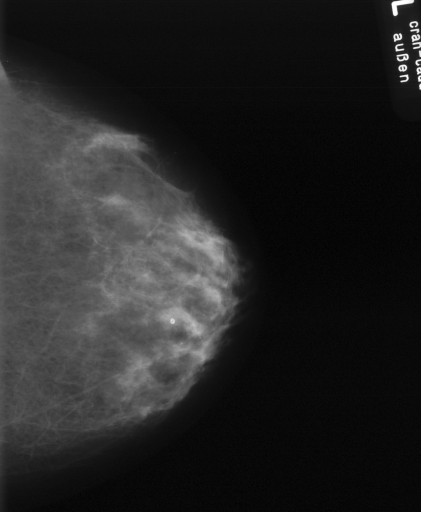}
\includegraphics[width=0.19\columnwidth,height=2cm]{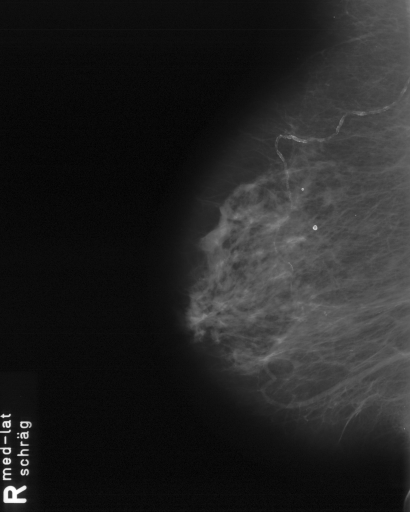}
\includegraphics[width=0.19\columnwidth,height=2cm]{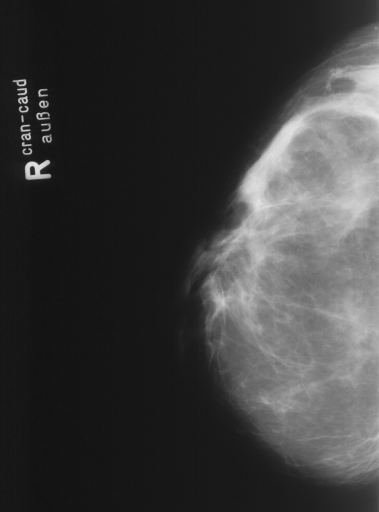}
\includegraphics[width=0.19\columnwidth,height=2cm]{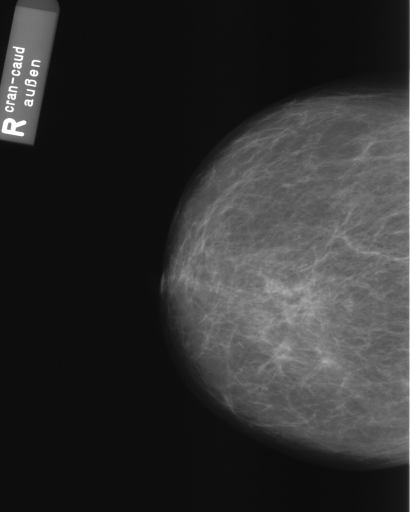}
\caption{Five breast images randomly selected from IRMA dataset (from left to right: b1, b2, b3, b4 and b5)}
\label{fig:breast}
\end{figure}

\begin{figure}[h]
\centering
\includegraphics[width=0.19\columnwidth,height=2cm]{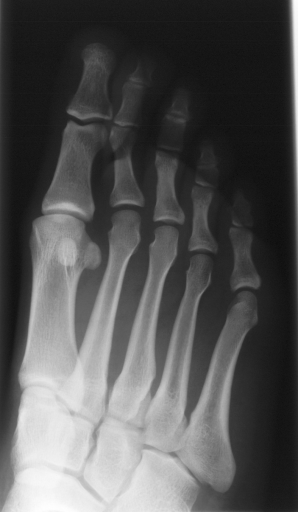}
\includegraphics[width=0.19\columnwidth,height=2cm]{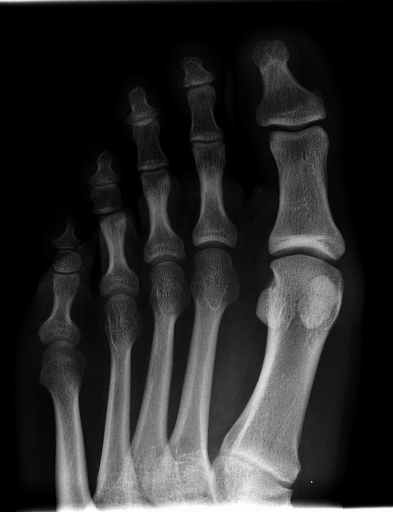}
\includegraphics[width=0.19\columnwidth,height=2cm]{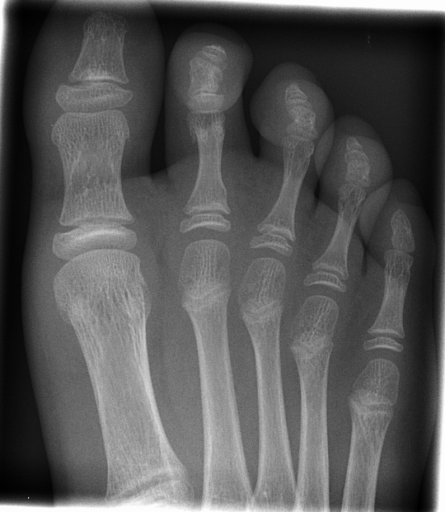}
\includegraphics[width=0.19\columnwidth,height=2cm]{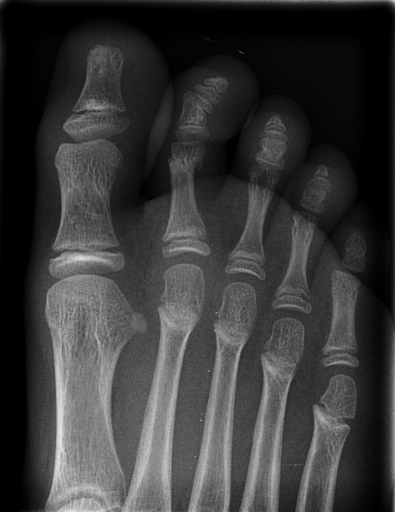}
\includegraphics[width=0.19\columnwidth,height=2cm]{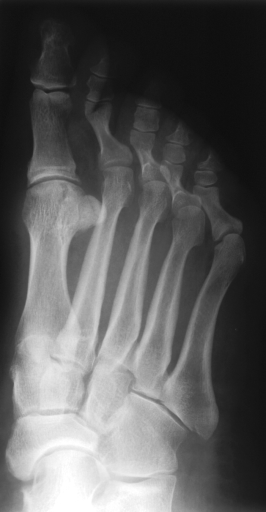}
\caption{Five foot x-ray images randomly selected from IRMA dataset (from left to right: f1, f2, f3, f4 and f5)}
\label{fig:foot}
\end{figure}

\begin{figure}[h]
\centering
\includegraphics[width=0.19\columnwidth,height=2cm]{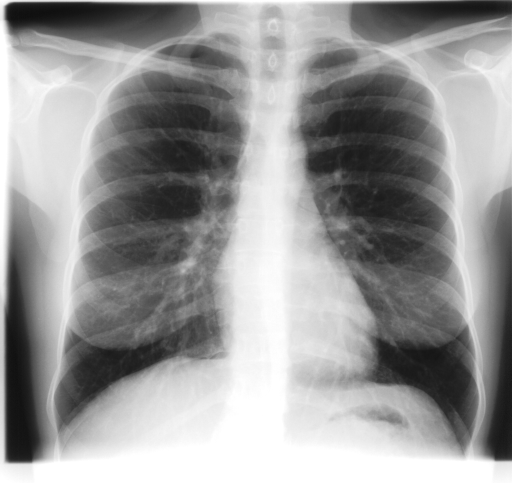}
\includegraphics[width=0.19\columnwidth,height=2cm]{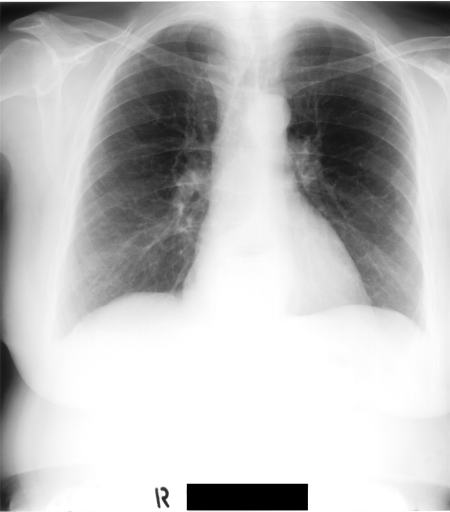}
\includegraphics[width=0.19\columnwidth,height=2cm]{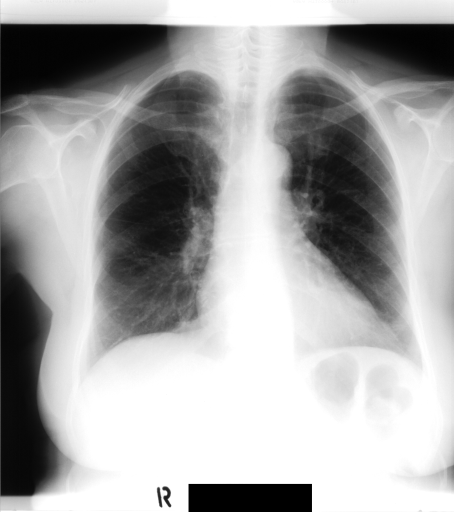}
\includegraphics[width=0.19\columnwidth,height=2cm]{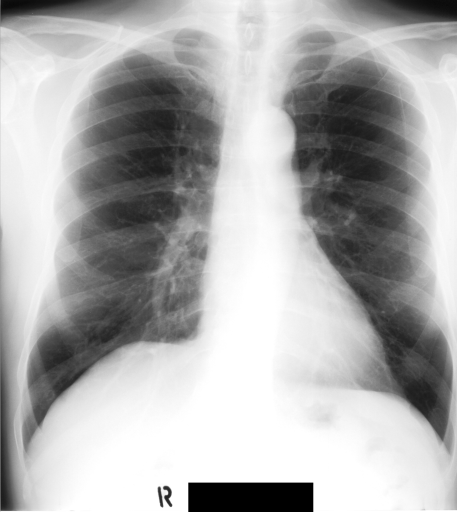}
\includegraphics[width=0.19\columnwidth,height=2cm]{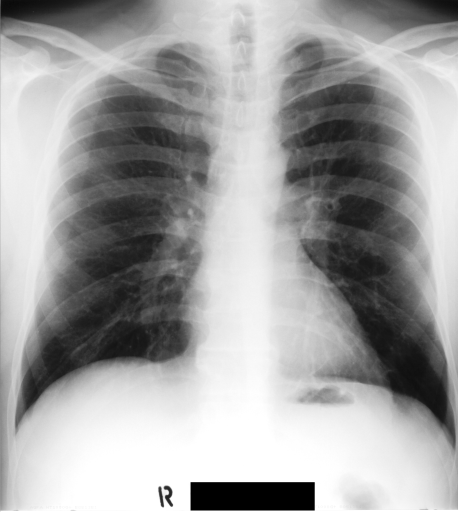}
\caption{Five chest x-ray images randomly selected from IRMA dataset (from left to right: l1, l2, l3, l4 and l5)}
\label{fig:lung}
\end{figure}

  In this series of experiments, we randomly selected three categories from IRMA dataset (namely, breast, foot and lung), and for each category we choose 5 random images from that class (Figures \ref{fig:breast}, \ref{fig:foot} and \ref{fig:lung} show these images). When we find ``4 out of 180'' (4/180) and ``8 out of 180'' (8/180) projections via micro-DE (tasks that cannot be performed via brute-force), then the question is what can we say about the reconstruction error, in terms of correlation between original and reconstructed images, when we compare evolutionary approximation with the case that we can manage via exhaustive search, namely ``4 out of 16'' (4/16)? Can micro-DE reach at least the same correlation (similarity) as the brute-force case in lower dimensions? If yes, we may have more confidence in using micro-DE in practice where we cannot apply brute force for benchmarking or direct use, especially for higher dimensions (i.e., the number of projections). 
 

For finding optimal projections based on micro-DE (4 out of 180), we set the parameters of micro-DE to be NFC$_{\max}=300$, $N_p=6$, $F=0.5$, $C_r=0.9$, and we run each experiment 30 times. For optimal projections based on micro-DE (8 out of 180), we set NFC$_{\max}=400$, $N_p=10$, $F=0.5$, $C_r=0.9$, and we run each experiment 30 times. The results are presented in Tables \ref{tab:breastRes}, \ref{tab:footRes} and \ref{tab:lungRes}. In all cases, micro-DE (MDE) with 4/180 reaches the same correlations as brute force (BF) for 4/16. Of course, MDE finds different projections as it is searching the entire search space of $180 \choose 4$. However, that the same level of reconstruction accuracy can be achieved establishes the reliability of MDE as a practical solution that may even produce a higher-level of uniqueness for Radon barcodes as 4 projections are selected among all 180 angles (the increased uniqueness needs to be verified by applying the Radon barcodes for image retrieval). On the other hand, MDE for 8/180 clearly increases the correlation with statistical significance. This is very encouraging as we can generate more expressive Radon barcodes using a higher number of projections.       

\begin{table*}[!t]
\centering
\caption{Results for five breast images from Figure \ref{fig:breast}.}
\label{tab:breastRes}
\begin{tabular}{l|ll|ll||ll}
image & BF (4/16) & $C_{\max}$ & MDE (4/180) & $C_{\max}$ & MDE (8/180) & $C_{\max}$ \\ \hline  
b1  & [22,67,123,157] & 0.83 & [30,50,120,160] & 0.83 & [70,130,30,120,100,150,170,50] & 0.89 \\
b2  & [33,67,123,169] & 0.83 & [30,70,120,160] & 0.83 & [170,140,150,60,40,80,20,110] & 0.89 \\
b3  & [22,67,123,157] & 0.81 & [30,50,120,160] & 0.81 & [150,170,10,70,40,130,110,50] & 0.85 \\
b4  & [11,56,146,169] & 0.81 & [20,60,120,160] & 0.80 & [160,20,120,170,40,30,130,70] & 0.86 \\
b5  & [11,56,112,157] & 0.81 & [20,60,120,160] & 0.81 & [80,20,120,60,160,90,10,140] & 0.86 \\ \hline     
\end{tabular}
\end{table*}
\begin{table*}[!t]
\centering
\caption{Results for five foot images from Figure \ref{fig:foot}.}
\label{tab:footRes}
\begin{tabular}{l|ll|ll||ll}
image & BF (4/16) & $C_{\max}$ & MDE (4/180) & $C_{\max}$ &  MDE (8/180) & $C_{\max}$ \\ \hline  
f1  & [34,67,112,157] & 0.79 & [30,80,110,150] & 0.80 & [80,160,30,130,30,100,140,60] & 0.85 \\
f2  & [79,101,135,169] & 0.73 & [80,110,140,170] & 0.73 & [70,20,160,0,120,80,130,100] & 0.80 \\
f3  & [11,56,112,158] & 0.64 & [20,60,100,170] & 0.65 & [40,130,160,70,10,170,110,60] & 0.70 \\
f4  & [11,79,101,169] & 0.77 & [0,30,80,100] & 0.77 & [20,180,50,60,80,110,170,120] & 0.84 \\
f5  & [22,67,112,157] & 0.85 & [80,30,120,160] & 0.86 & [120,60,80,10,150,100,170,40] & 0.89 \\ \hline     
\end{tabular}
\end{table*}
\begin{table*}[!t]
\centering
\caption{Results for five lung images from Figure \ref{fig:lung}.}
\label{tab:lungRes}
\begin{tabular}{l|ll|ll||ll}
image & BF (4/16) & $C_{\max}$ & MDE (4/180) & $C_{\max}$ & MDE (8/180) & $C_{\max}$ \\ \hline  
f1  & [11,45,135,169] & 0.61 & [20,40,140,170] & 0.59 & [180,170,30,90,10,90,40,120] & 0.65 \\
f2  & [11,67,112,169] & 0.66 & [30,170,70,110] & 0.66 & [70,70,100,140,20,20,160,170] & 0.70 \\
f3  & [11,67,112,169] & 0.69 & [10,30,120,170] & 0.68 & [30,50,100,110,80,10,150,180] & 0.72 \\
f4  & [22,78,112,157] & 0.59 & [20,80,120,160] & 0.60 & [70,40,10,150,180,90,50,130] & 0.66 \\
f5  & [22,78,101,169] & 0.58 & [20,90,130,170] & 0.61 & [110,140,40,20,90,80,170,10] & 0.65 \\ \hline     
\end{tabular}
\end{table*}

\begin{figure}[htb]
\centering
\vspace{0.05in}
\includegraphics[width=0.79\columnwidth]{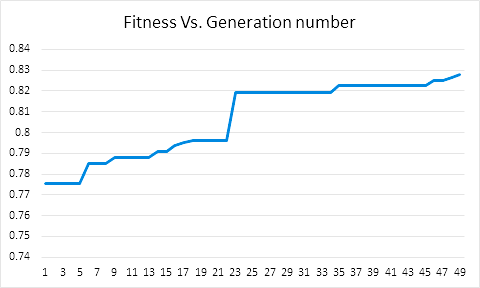}\vspace{0.01in}
\includegraphics[width=0.79\columnwidth]{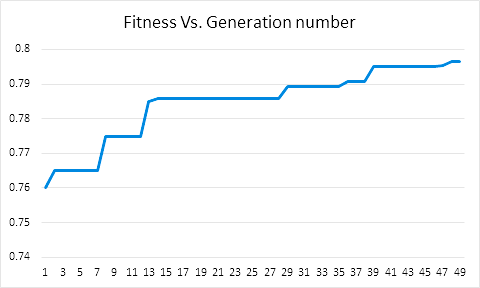}\vspace{0.01in}
\includegraphics[width=0.79\columnwidth]{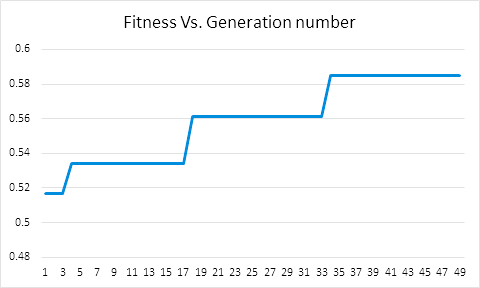}
\caption{Fitness curves for b1, f1 and l1.}
\label{fig:fitnessDiag}
\end{figure}

Table \ref{tab:allResults} show the average correlation (accuracy) of brute force versus micro-DE for all 10 image classes. As it can be observed micro-DE can find optimal angles for 4/180 case with statistically the same correlation as for brute force in case of 4/16. Further, micro-DE can achieve higher accuracies in case we configure it for 8/180. We run experiments in Matlab (R2015a) on a Windows 8 machine (LENEVO, Intel i7, 2.6GHz with 8GB memory). The brute force approach (4/16) took on average 16.51 seconds fro each image. Micro-DE needed 8.23 seconds for 4/16, 10.90 seconds for 4/180, and 16.28 seconds for 8/180.

\begin{table}[tb]
\centering
\caption{Average correlations for all 10 classes (the image in first column is representative for the classes from Figure \ref{fig:sampleXrays}). }
\label{tab:allResults}
\begin{tabular}{l||l||l|l}
class & $\mu_{C_{BF,4/16}}$ & $\mu_{C_{MDE,4/180}}$ &$\mu_{C_{MDE,8/180}}$ \\ \hline  
\includegraphics[width=0.22in,height=0.22in]{b1.png}    & $0.82\pm 0.01$	&  $0.82\pm 0.01$ &  $0.87\pm 0.02$ \\
\includegraphics[width=0.22in,height=0.22in]{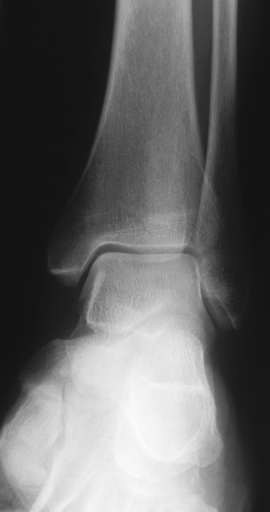}    & $0.87\pm 0.01$ &	 $ 0.86 \pm 0.01$ & $0.89\pm 0.01$\\
\includegraphics[width=0.22in,height=0.22in]{f1.png} & $0.76\pm 0.08$ & 	$0.76\pm 0.08$ & $0.82\pm 0.07$   \\
\includegraphics[width=0.22in,height=0.22in]{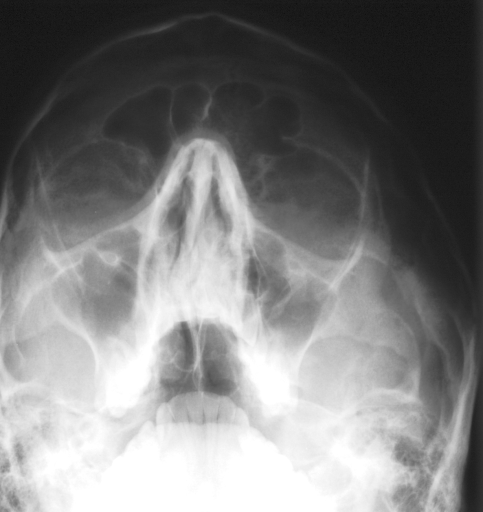} & $0.78\pm 0.06$ & 	$0.78\pm 0.06$ & $0.82\pm 0.05$   \\
\includegraphics[width=0.22in,height=0.22in]{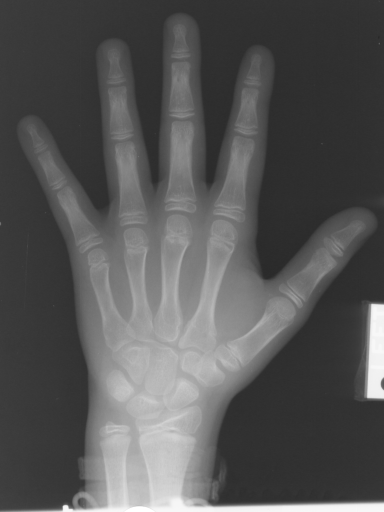} &   $0.76\pm 0.06$ & 	  $0.77\pm 0.05$ & $0.81\pm 0.06$   \\
\includegraphics[width=0.22in,height=0.22in]{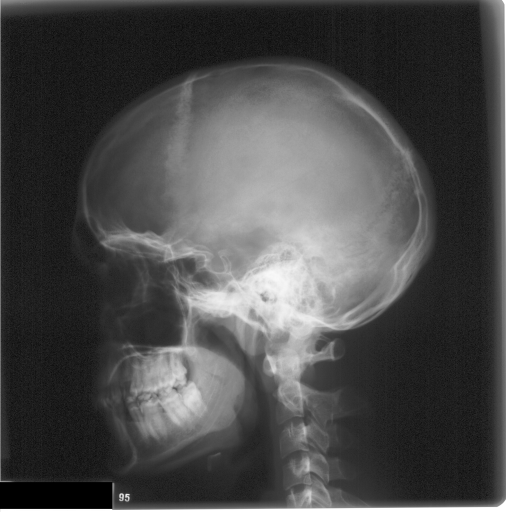} & $0.83\pm 0.04$ &	$0.83 \pm 0.04$ & $0.87\pm 0.05$    \\
\includegraphics[width=0.22in,height=0.22in]{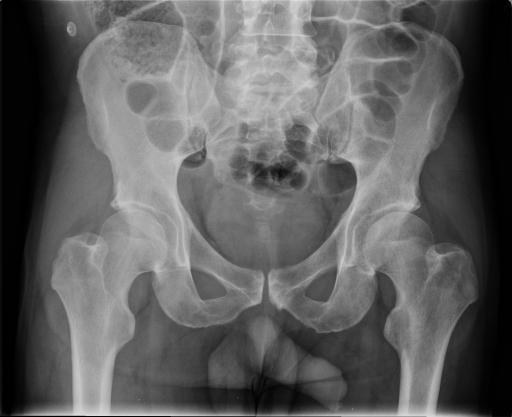} & $0.80\pm 0.09$ &  	$0.80\pm 0.09$ & $0.84\pm 0.09$   \\
\includegraphics[width=0.22in,height=0.22in]{l1.png} & $0.63\pm 0.05 $ & 	$0.63\pm 0.04$ & $0.68\pm 0.03$   \\
\includegraphics[width=0.22in,height=0.22in]{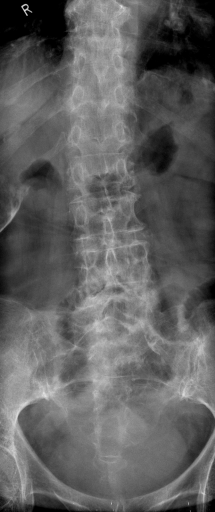} &  $0.67\pm 0.09$ & 	$0.67\pm 0.08$ & $0.71\pm 0.08$  \\
\includegraphics[width=0.22in,height=0.22in]{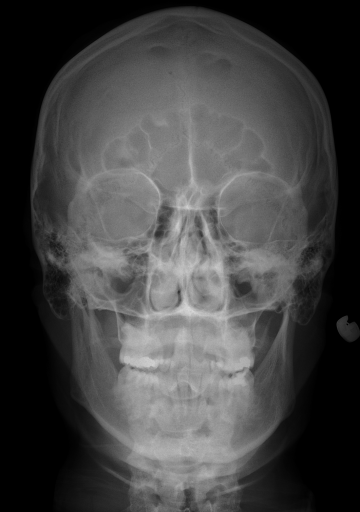} & $0.92\pm 0.01 $ & 	$0.91\pm 0.01$	& $0.94\pm 0.02$ \\
\end{tabular}
\end{table}

\section{Conclusions}
\label{sec:conc}
 
 Image retrieval from a large archive of digital images is a necessary and challenging task. More and more researchers have investigated the feasibility of binary descriptors to tag images such that they can be found and retrieved easier/faster. In medical imaging this may have tremendous positive impact on diagnostic accuracy in both radiology and pathology; by examining similar cases, the expert can make more reliable decisions. One of the recent attempts to generate binary tags is the generation of ``Radon barcodes'' where equidistant projection angles are binarized and concatenated to assemble a barcode for image search. This seems to be particularly useful for medical images. 

In this paper, we proposed an evolutionary approach to locating an optimal subset of projection angles (e.g., 4 or 8 projection angles out of all 180 angles) in order to increase the expressiveness of Radon barcodes. Preliminary results show that this approach may contribute to the uniqueness, and hence to higher retrieval accuracy of Radon barcodes. In future works, the proposed scheme needs to be integrated with a complete retrieval system to verify its overall benefit for content-based medical image retrieval.

\bibliographystyle{IEEEbib}
\bibliography{refs}

\end{document}